\pdfoutput=1

\documentclass[sigconf,natbib=true,anonymous=false,review=false]{acmart}
\usepackage{soul}
\usepackage{threeparttable}
\usepackage{multirow}
\usepackage{subcaption}
\usepackage{graphicx}
\usepackage{xcolor}

\definecolor{VibrantBlue}{rgb}{0.0, 0.2, 1.0}

\copyrightyear{2025}
\acmYear{2025}
\setcopyright{cc}
\setcctype{by}
\acmConference[WWW Companion '25]{Companion Proceedings of the ACM Web Conference 2025}{April 28-May 2, 2025}{Sydney, NSW, Australia}
\acmBooktitle{Companion Proceedings of the ACM Web Conference 2025 (WWW Companion '25), April 28-May 2, 2025, Sydney, NSW, Australia}
\acmDOI{XX.XXXX/XXXXXXX.XXXXXXX}
\acmISBN{979-8-4007-1331-6/25/04}

\begin{document}
\title{Mapping and Influencing the Political Ideology of Large Language Models using Synthetic Personas}

\author{Pietro Bernardelle}
\orcid{0009-0003-3657-9229} 
\affiliation{%
  \institution{The University of Queensland}
  \city{Brisbane}
  \country{Australia}
}
\email{p.bernardelle@uq.edu.au}

\author{Leon Fröhling}
\orcid{0000-0002-5339-7019} 
\affiliation{%
  \institution{GESIS}
  \city{Cologne}
  \country{Germany}
}
\email{leon.froehling@gesis.org}

\author{Stefano Civelli}
\orcid{0009-0003-4982-9565} 
\affiliation{%
  \institution{The University of Queensland}
  \city{Brisbane}
  \country{Australia}
}
\email{s.civelli@uq.edu.au}

\author{Riccardo Lunardi}
\orcid{0009-0001-5550-317X} 
\affiliation{%
  \institution{University of Udine}
  \city{Udine}
  \country{Italy}
}
\email{riccardo.lunardi@uniud.it}

\author{Kevin Roitero}
\orcid{0000-0002-9191-3280} 
\affiliation{%
  \institution{University of Udine}
  \city{Udine}
  \country{Italy}
}
\email{kevin.roitero@uniud.it}

\author{Gianluca Demartini}
\orcid{0000-0002-7311-3693} 
\affiliation{%
  \institution{The University of Queensland}
  \city{Brisbane}
  \country{Australia}
}
\email{demartini@acm.org}

\renewcommand{\shortauthors}{Pietro Bernardelle et al.}

\begin{abstract}
The analysis of political biases in large language models (LLMs) has primarily examined these systems as single entities with fixed viewpoints. While various methods exist for measuring such biases, the impact of persona-based prompting on LLMs' political orientation remains unexplored. In this work we leverage PersonaHub, a collection of synthetic persona descriptions, to map the political distribution of persona-based prompted LLMs using the Political Compass Test (PCT). We then examine whether these initial compass distributions can be manipulated through explicit ideological prompting towards diametrically opposed political orientations: right-authoritarian and left-libertarian. Our experiments reveal that synthetic personas predominantly cluster in the left-libertarian quadrant, with models demonstrating varying degrees of responsiveness when prompted with explicit ideological descriptors. While all models demonstrate significant shifts towards right-authoritarian positions, they exhibit more limited shifts towards left-libertarian positions, suggesting an asymmetric response to ideological manipulation that may reflect inherent biases in model training.
\end{abstract}



\maketitle
\newpage

\begin{figure}[ht]
    \centering
    \begin{subfigure}[b]{0.48\linewidth}
        \centering
        \includegraphics[width=\textwidth]{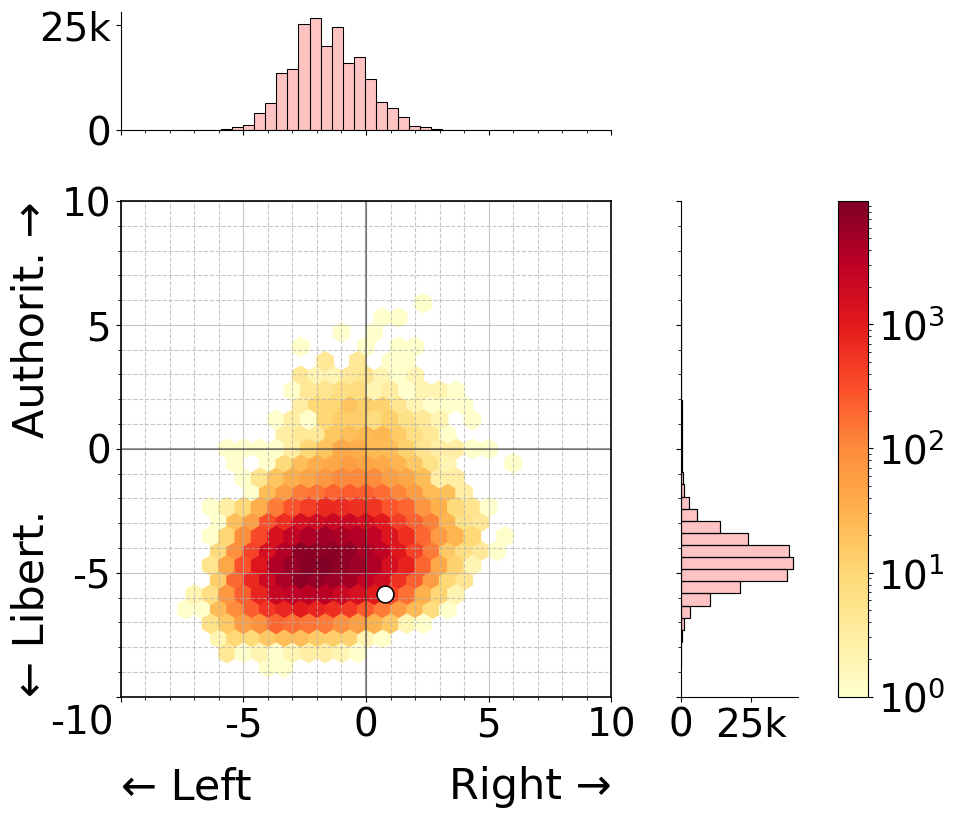}
        \caption{Mistral-7B-Instruct-v0.3}
        \label{fig:subfig1}
    \end{subfigure}
    \hfill
    \begin{subfigure}[b]{0.48\linewidth}
        \centering
        \includegraphics[width=\textwidth]{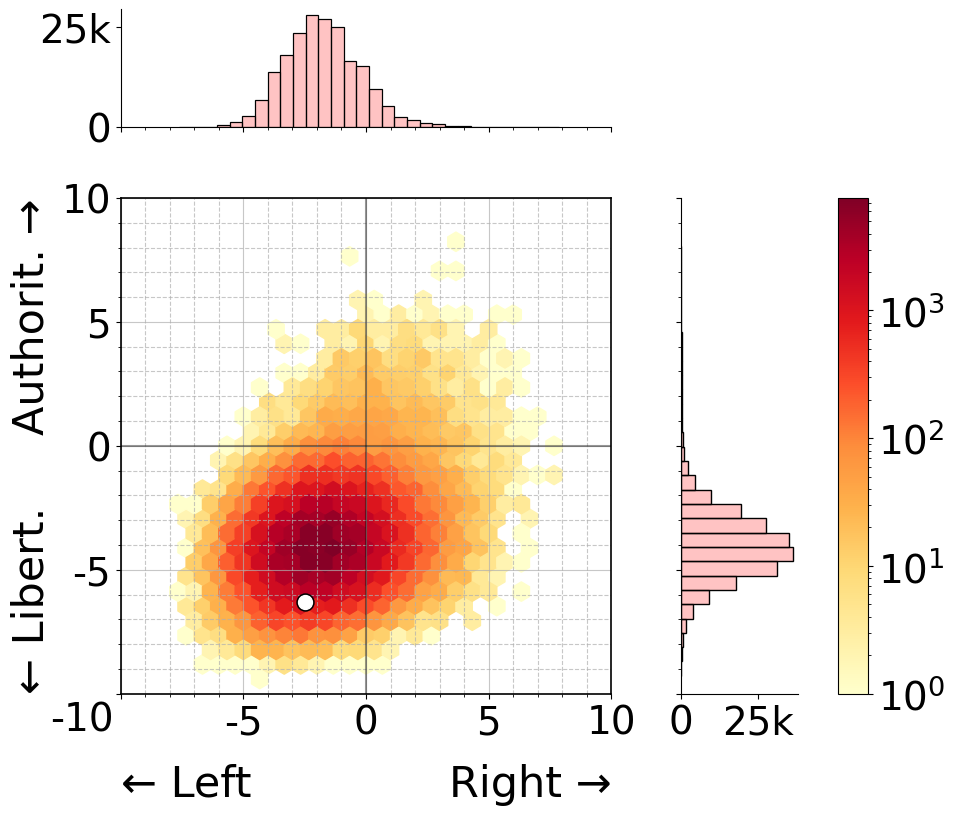}
        \caption{Llama-3.1-8B-Instruct}
        \label{fig:subfig2}
    \end{subfigure}
    
    \begin{subfigure}[b]{0.48\linewidth}
        \centering
        \includegraphics[width=\textwidth]{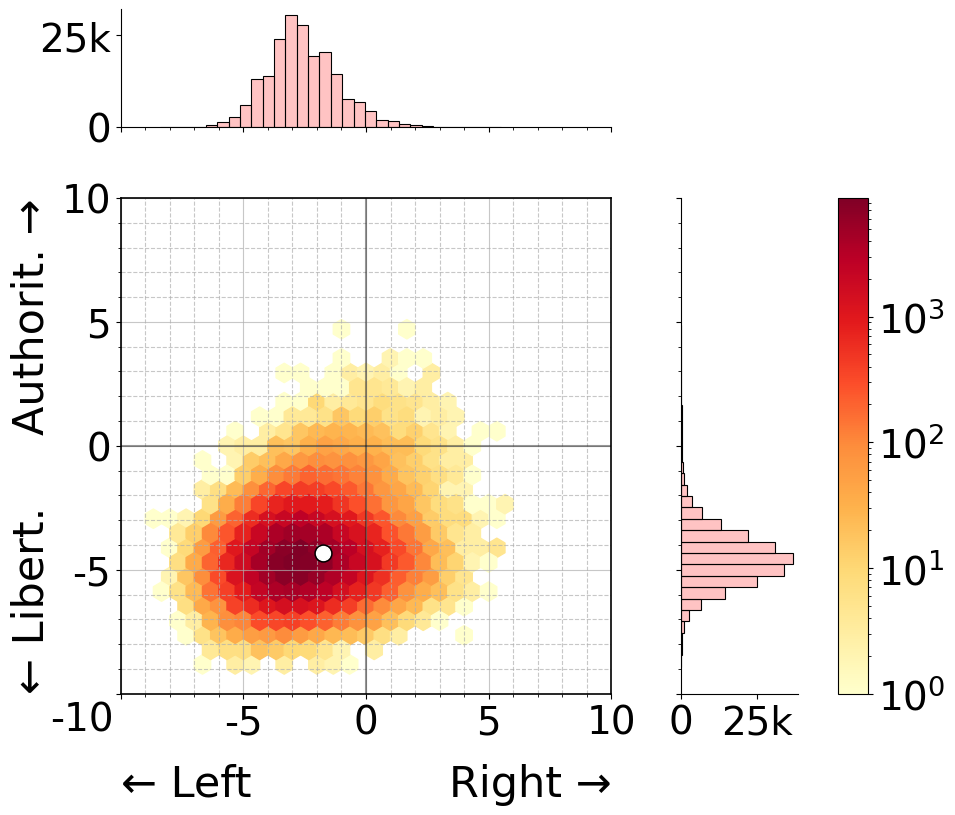}
        \caption{Qwen2.5-7B-Instruct}
        \label{fig:subfig3}
    \end{subfigure}
    \hfill
    \begin{subfigure}[b]{0.48\linewidth}
        \centering
        \includegraphics[width=\textwidth]{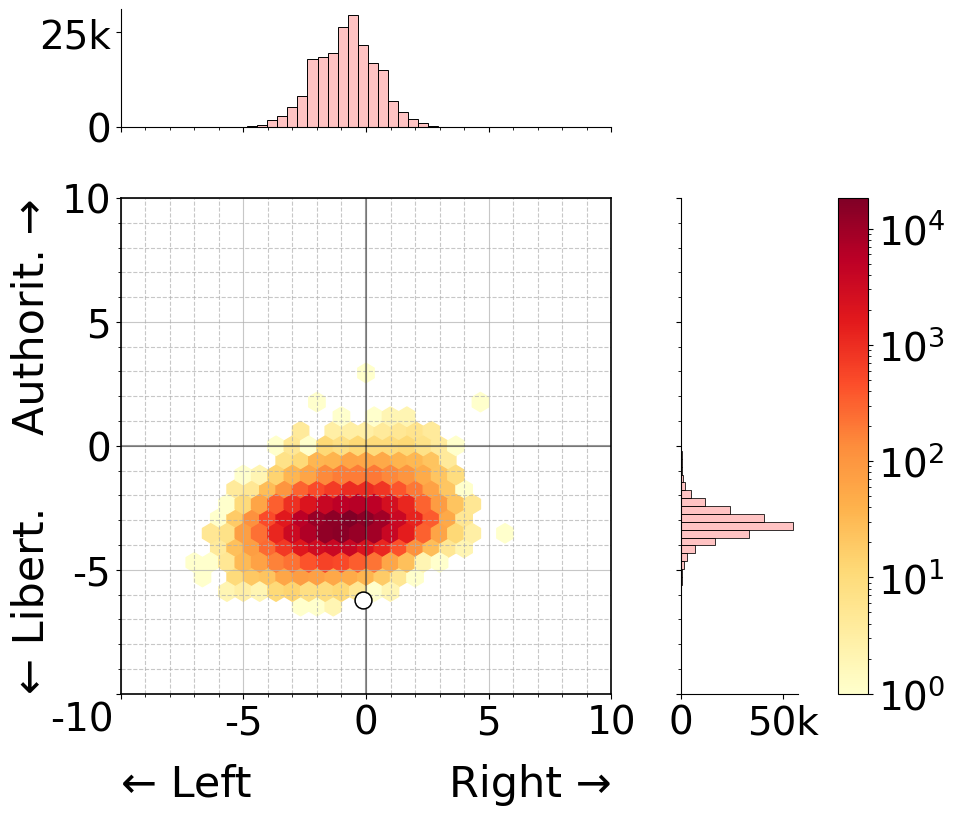}
        \caption{Zephyr-7b-beta}
        \label{fig:subfig4}
    \end{subfigure}
    
    \caption{Political compass distribution of PersonaHub personas when impersonated by different LLMs. Darker regions indicate higher density of personas on a logarithmic scale. The white dot represents the leaning of the original LLM (without any form of persona prompting). The bar charts show the marginal distributions along each axis.}
    \label{fig:combined}
\end{figure}

\section{Introduction}
Large Language Models (LLMs) have been shown to exhibit distinct political biases when evaluated through the Political Compass Test (PCT) \cite{feng-etal-2023-pretraining,rottger-etal-2024-political}, typically leaning towards economically left and socially liberal positions. Parallel research has demonstrated that these models can dynamically adopt the perspective of different personas through zero-shot prompting, increasing diversity in their outputs and behavior patterns \cite{frohling2024personas} while accurately simulating responses across demographic groups \cite{Argyle_2023}. However, this adaptability comes with potential drawbacks, as it can amplify stereotypes \cite{liu2024evaluating} and significantly alter political expression \cite{bang2024measuring}. This malleability in LLMs' behavior raises questions about the stability of their political orientations and the potential for deliberate manipulation through persona-based prompting.

Recent work has introduced PersonaHub \cite{ge2024scaling}, a collection of synthetic persona descriptions generated through LLM bootstrapping. While these personas have been shown to increase diversity in various natural language processing (NLP) tasks, their influence on LLMs' political biases remains unexplored. 
Our study bridges this gap by investigating how different PersonaHub personas affect LLMs' responses to the PCT. Specifically, we examine both the distribution of political orientations across personas and the potential for manipulation through explicit ideological prompting.
We focus on testing ideological malleability towards right-authoritarian and left-libertarian positions, as these represent maximally distant points on the political compass while also allowing us to examine both reinforcement and resistance to LLMs' inherent left-libertarian bias. Our investigation addresses two questions:\vspace{1mm}

\begin{itemize}
    \item[\textbf{RQ1:}] How do diverse persona descriptions affect LLMs' positions on the Political Compass Test?
    \item[\textbf{RQ2:}] To what extent the introduction of explicit ideological elements in persona descriptions can influence LLMs' Political Compass Test outcomes?
\end{itemize}

We show that synthetic personas predominantly cluster in the left-libertarian quadrant (Figure \ref{fig:combined}), with Mistral, Llama, and Qwen exhibiting similar distributions. Zephyr demonstrates the highest concentration of personas around its centroid, while Llama's personas distribution displays the greatest variance.
Additionally, explicit ideological prompting can substantially alter this baseline distribution, though models respond differently to such interventions. In our experiments, Llama exhibited the strongest shift toward right-authoritarian positions, while Mistral showed the most pronounced movement toward left-libertarian orientations.

\section{RELATED WORK}
\paragraph{\textbf{Personas in LLMs}}
Recent work has shown that LLMs can effectively adopt and simulate different personas through appropriate prompting. PersonaHub \cite{ge2024scaling} introduced a collection of over 1 billion synthetically generated persona descriptions, showing that these personas can help diversify model outputs in a controlled way. This approach has been successful in diversifying data annotations, where personas help elicit a wider range of valid perspectives \cite{frohling2024personas}. The ability of LLMs to reliably simulate different personas has also been explored in the context of simulating human samples \cite{Argyle_2023}, where LLMs successfully predicted survey responses across different demographic groups with high accuracy.
Our work builds on these findings by systematically mapping personas from PersonaHub onto the political compass, providing the first large-scale analysis of how synthetic personas distribute across political ideological space over multiple LLMs.

\paragraph{\textbf{LLMs \& Political Compass Test}}
Multiple studies have investigated the political biases of LLMs using the PCT as an evaluation framework. Initial work by \citet{hartmann2023political} found that ChatGPT exhibited consistently left-libertarian leanings. Subsequent studies confirmed this tendency across different LLMs while highlighting challenges in reliably eliciting political stances from more recent models with stronger guardrails \cite{rottger-etal-2024-political}. Recent methodological innovations have focused on making PCT evaluations more robust, including through masked token prediction \cite{feng-etal-2023-pretraining} and ensuring stability across different prompting approaches \cite{rottger-etal-2024-political}. Our work advances this line of research by combining PCT evaluation with persona-based prompting, offering a novel methodology for understanding not just the default political biases of LLMs, but also their capacity to simulate diverse political perspectives through persona adoption.

\begin{table}[t]
\centering
\small
\caption{Average distance between each LLM-generated persona response and the group's central point, measured across three political configurations: baseline (plain description), right-authoritarian descriptor, and left-libertarian descriptor. Lower values indicate tighter clustering of persona responses within the political compass coordinate system.}
\label{tab:avg-results}
\begin{tabular}{lccc}
\toprule
\multirow{2}{*}{Model} & \multicolumn{3}{c}{Avg Distance} \\
\cmidrule(lr){2-4}
& Base & Right-Authoritarian & Left-Libertarian \\
\midrule
Mistral & 1.518 & 1.766 & 0.950 \\
Llama & 1.758 & 1.801 & 1.606 \\
Qwen2.5 & 1.560 & 1.348 & 0.833 \\
Zephyr & 1.164 & 1.175 & 1.163 \\
\bottomrule
\end{tabular}
\end{table}

\section{METHODOLOGY}
\paragraph{\textbf{Data}} We utilize PersonaHub, a collection of 1 billion diverse personas\footnote{with 200,000 publicly accessible through Huggingface}, to examine how persona descriptions shape the political orientations of LLMs. To systematically evaluate these influences, we employ the PCT, which assesses political perspectives through 62 distinct statements across six key domains. While previous research by \citet{kevin} has shown that LLMs' responses may vary based on question phrasing, we maintain the original PCT formulations to ensure methodological consistency. Our analysis includes 12.4 million categorical responses (200,000 personas x 62 statements), collected by prompting models to select from predefined political stance options.

\paragraph{\textbf{Models}} For our analysis, we select four open-source language models: Mistral-7B-Instruct-v0.3, Llama-3.1-8B-Instruct, Qwen-2.5-7B, and Zephyr-7B-beta. These models were chosen for their open-source nature and relatively small parameter size (7-8B), enabling easy reproducibility while maintaining sufficient diversity for generalizable insights.

\paragraph{\textbf{Experimental setup}} Our experimental setup comprises of two phases. First, we establish a baseline by having each model complete the PCT while impersonating each persona description. Second, we investigate ideological malleability by injecting explicit political descriptors (``right authoritarian" or ``left libertarian") to the persona descriptions and measuring the shifts in their political positioning.\footnote{Code and data are openly available at \url{https://github.com/d-lab/llm-political-personas}}

\begin{figure*}[t]
    \centering
    \begin{subfigure}[b]{0.23\linewidth}
        \centering
        \includegraphics[width=\textwidth]{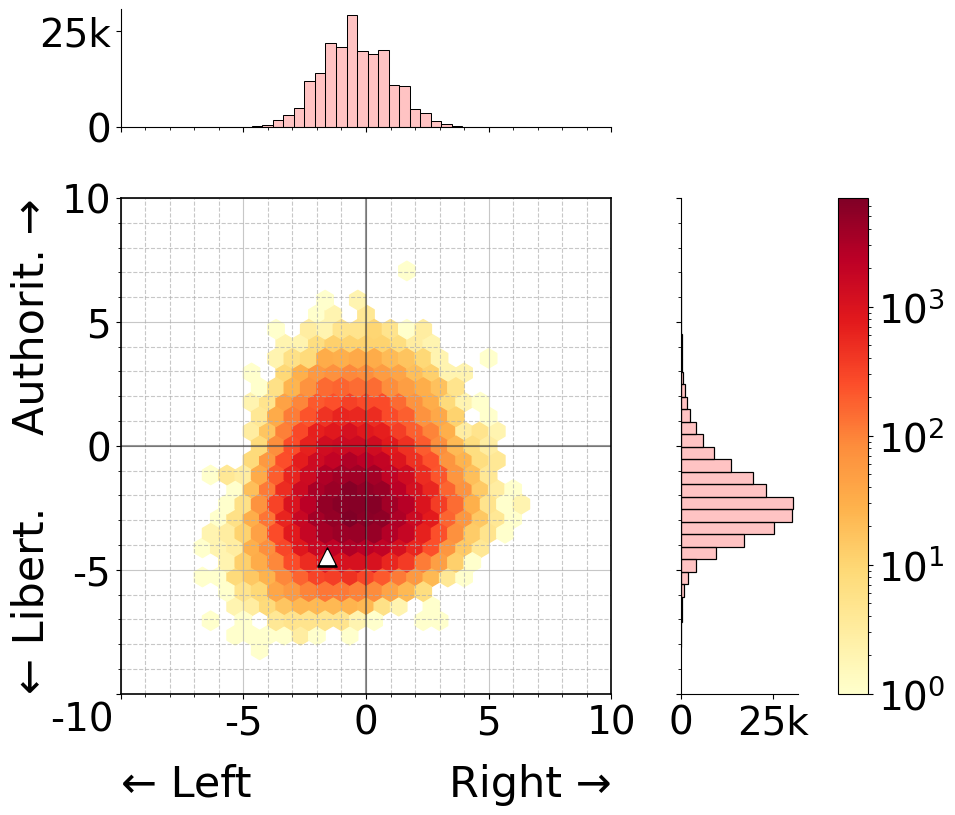}
        \label{fig:subfig1_righta}
    \end{subfigure}
    \hfill
    \begin{subfigure}[b]{0.23\linewidth}
        \centering
        \includegraphics[width=\textwidth]{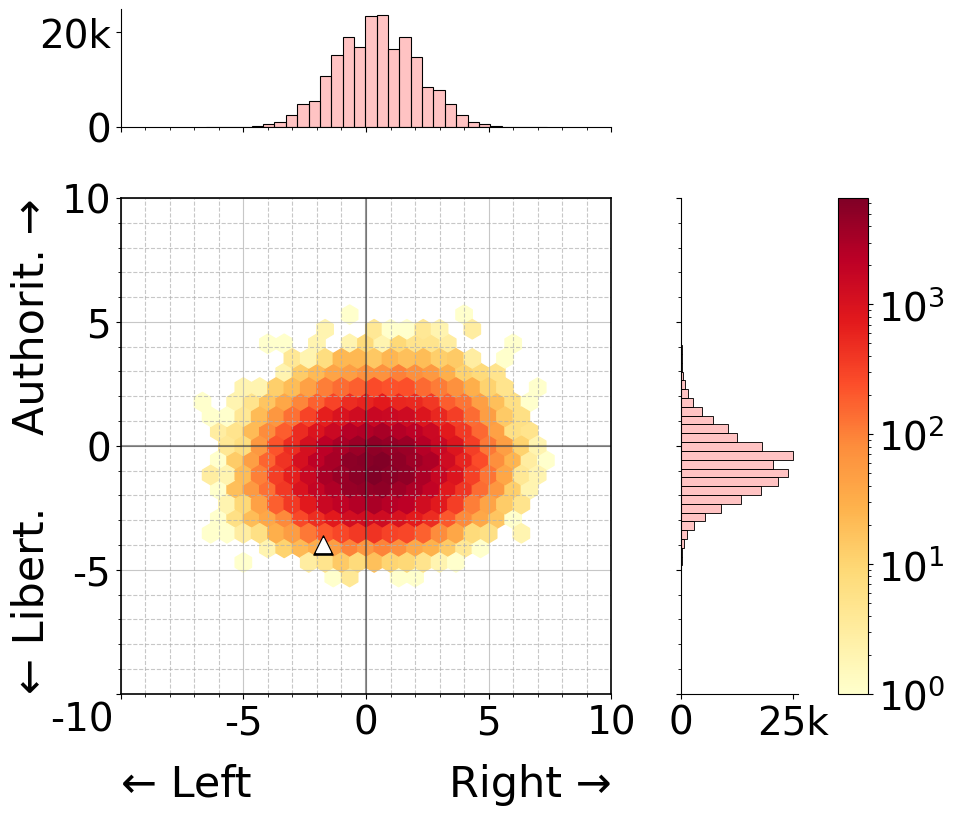}
        \label{fig:subfig2_righta}
    \end{subfigure}
    \hfill
    \begin{subfigure}[b]{0.23\linewidth}
        \centering
        \includegraphics[width=\textwidth]{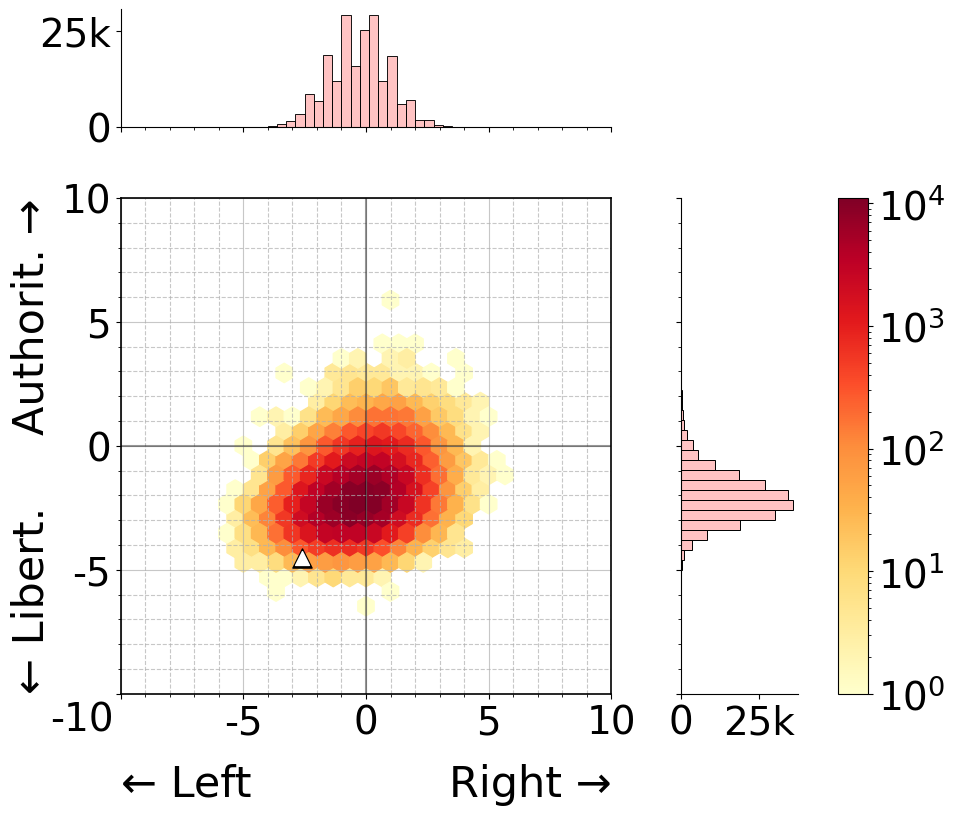}
        \label{fig:subfig3_righta}
    \end{subfigure}
    \hfill
    \begin{subfigure}[b]{0.23\linewidth}
        \centering
        \includegraphics[width=\textwidth]{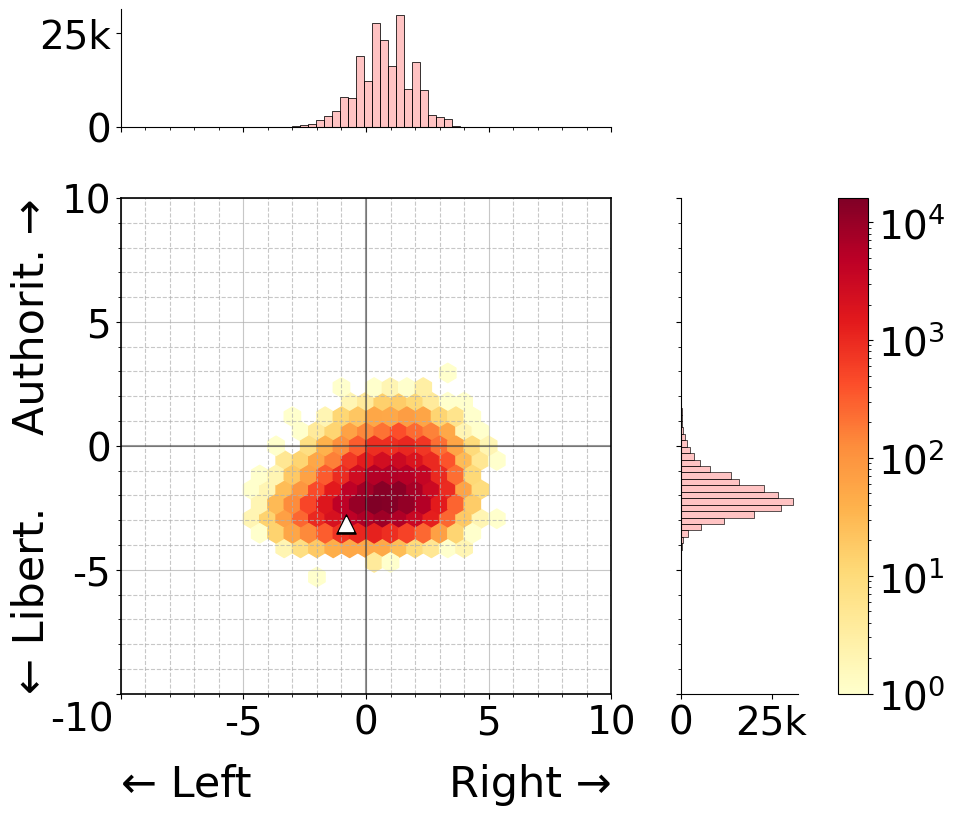}
        \label{fig:subfig4_righta}
    \end{subfigure}
    
    \begin{subfigure}[b]{0.23\linewidth}
        \centering
        \includegraphics[width=\textwidth]{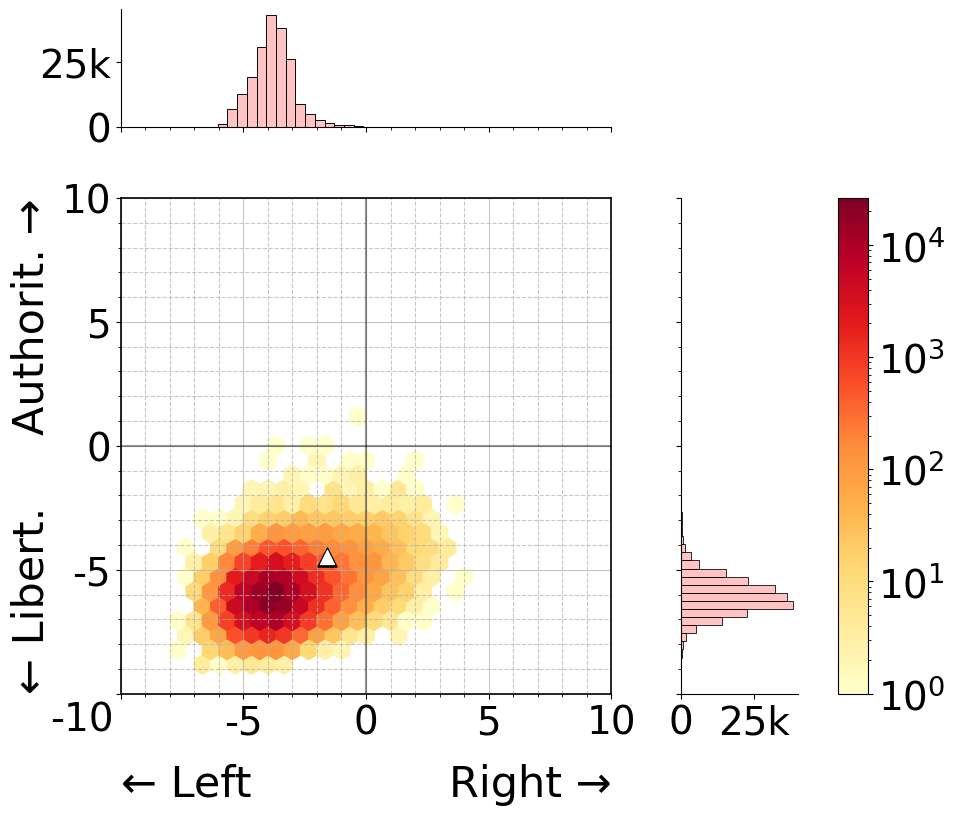}
        \caption{Mistral-7B-Instruct-v0.3}
        \label{fig:subfig1_leftlib}
    \end{subfigure}
    \hfill
    \begin{subfigure}[b]{0.23\linewidth}
        \centering
        \includegraphics[width=\textwidth]{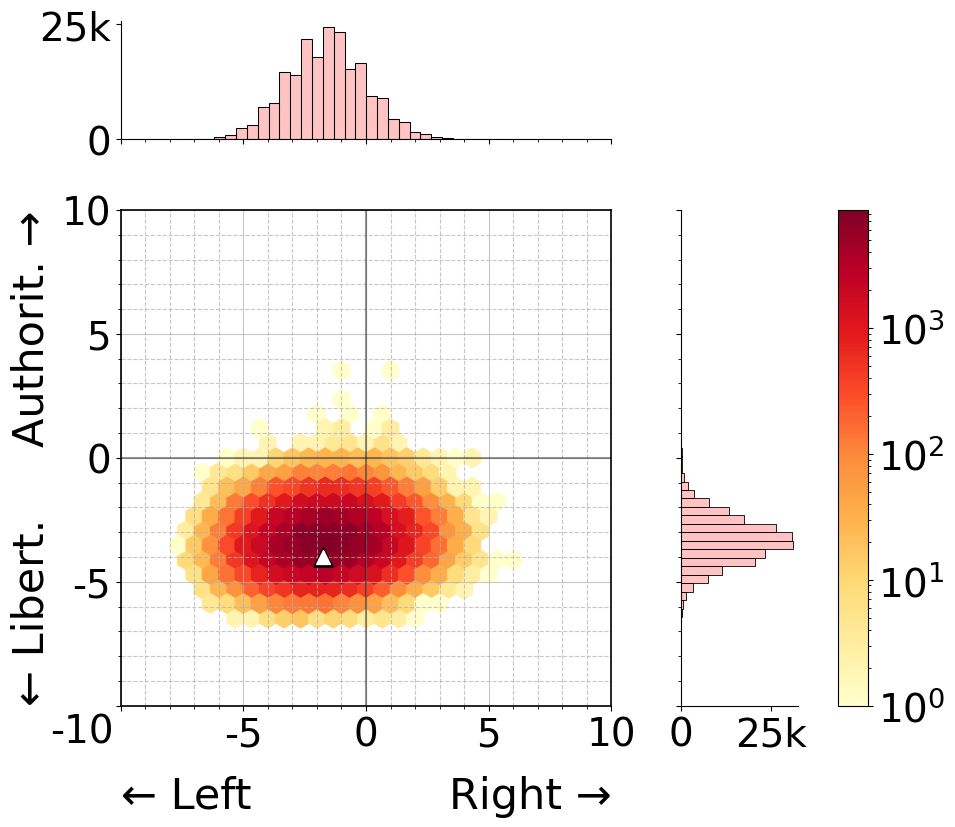}
        \caption{Llama-3.1-8B-Instruct}
        \label{fig:subfig2_leftlib}
    \end{subfigure}
    \hfill
    \begin{subfigure}[b]{0.23\linewidth}
        \centering
        \includegraphics[width=\textwidth]{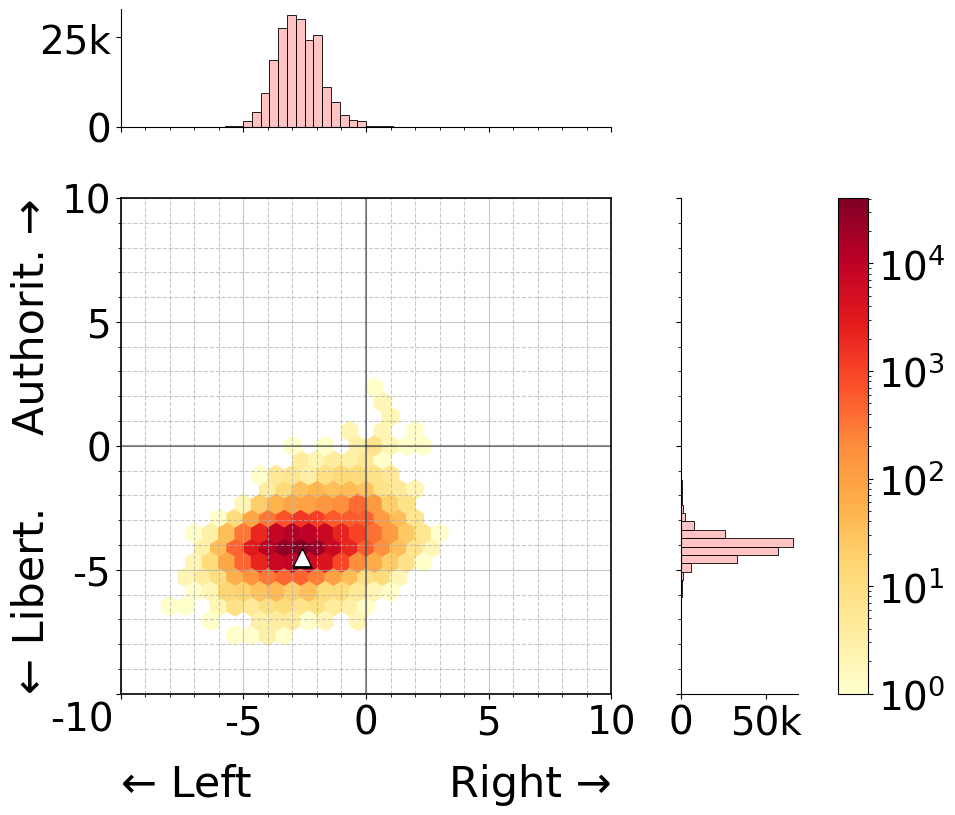}
        \caption{Qwen2.5-7B-Instruct}
        \label{fig:subfig3_leftlib}
    \end{subfigure}
    \hfill
    \begin{subfigure}[b]{0.23\linewidth}
        \centering
        \includegraphics[width=\textwidth]{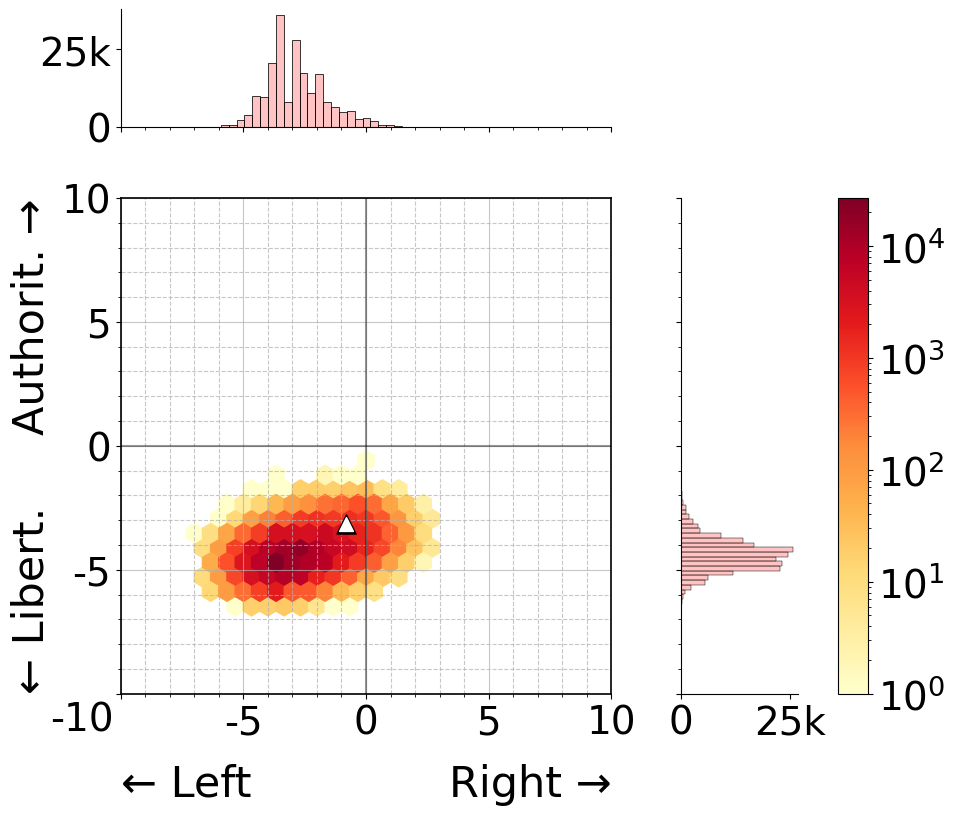}
        \caption{Zephyr-7b-beta}
        \label{fig:subfig4_leftlib}
    \end{subfigure}
    \caption{Political compass distribution of PersonaHub personas when impersonated by different LLMs. Top: Distribution after injecting the ``right-authoritarian" descriptor. Bottom: Distribution after injecting the ``left-libertarian" descriptor. Darker regions indicate higher density of personas on a logarithmic scale. The white triangle shows the average political position of the model across all persona-based prompts without explicit descriptor injection (i.e., the centroid from Figure \ref{fig:combined}).}
    \label{fig:political-compass}
\end{figure*}

\paragraph{\textbf{Computational resources}}
For our experiments, we used 8$\times$A100 GPU cards. The inference time was approximately four hours to run the PCT across each persona-based LLM, with a total experimental runtime of approximately 36 hours for the complete set of experiments.

\section{RESULTS}
\paragraph{\textbf{Personas political distribution}}
The baseline political distribution of personas, shown in Figure 1, reveals several patterns on how LLMs inherently position them in the political compass. Mistral, Llama, and Qwen exhibit similar distributions, with the majority of personas clustering in the left-libertarian quadrant—a reflection of the widely recognized left-libertarian bias encoded in LLMs.
While Zephyr maintains this left-libertarian tendency, its distribution is notably more concentrated (as shown in Table \ref{tab:avg-results}), and positioned closer to the center of the compass compared to the other three models. This tighter clustering and more moderate positioning suggests that Zephyr either has stronger internal constraints on ideological expression or is less responsive to the political variations implied by different personas.
Notably, while the personas distribution remain remarkably consistent across models, they do not align with the models' default positions (white dots in Figure \ref{fig:combined}). This difference between models' default political positions and their persona-based stances indicates that using personas may trigger consistent political behavioral patterns, regardless of the models' initial biases.
While the majority of personas align with left-libertarian tendencies, all four models demonstrate a minor presence in the right-authoritarian quadrant. This deviation from default positions indicates that, despite their biases, the models are capable of associating personas with politically diverse descriptors.

\paragraph{\textbf{Personas manipulation}}

\begin{table*}[t]

\centering
\small

\caption{Statistical analysis of LLMs' shifts under ``right-authoritarian" and ``left-libertarian" injection.}

\label{tab:model-stats-combined}

\begin{tabular}{lccccccccc}

\toprule

Condition & Model & $\Delta\mu$ ($\sigma$) & WSR z-score & d & 95\% CI & $\Delta\mu$ ($\sigma$) & WSR z-score & d & 95\% CI \\

\cmidrule(lr){1-10}

& & \multicolumn{4}{c}{X Variable} & \multicolumn{4}{c}{Y Variable} \\

\cmidrule(lr){3-6} \cmidrule(lr){7-10}

Right-authoritarian 
& Mistral & 1.18 (1.82) & -257.71*** & 0.84 & [0.83, 0.85] & 2.26 (1.44) & -382.29*** & 1.81 & [1.80, 1.82] \\ 
& Llama & 2.19 (2.20) & -326.95*** & 1.37 & [1.36, 1.38] & 3.20 (1.60) & -384.14*** & 2.51 & [2.50, 2.52] \\ 
& Qwen & 2.29 (1.80) & -359.97*** & 1.72 & [1.71, 1.73] & 2.44 (1.27) & -384.48*** & 2.37 & [2.36, 2.38] \\ 
& Zephyr & 1.60 (1.55) & -337.63*** & 1.36 & [1.35, 1.37] & 1.13 (0.88) & -371.57*** & 1.65 & [1.64, 1.66] \\ 

\cmidrule(lr){1-10}

Left-libertarian 
& Mistral & -2.18 (1.51) & -375.51*** & -1.82 & [-1.83, -1.81] & -1.57 (0.99) & -380.45*** & -1.81 & [-1.82, -1.80] \\ 
& Llama & 0.16 (2.13) & -51.09*** & 0.10 & [0.09, 0. 11] & 0.68 (1.52) & -199.00*** & 0.59 & [0.58, 0.60] \\ 
& Qwen & -0.15 (1.56) & -57.88*** & -0.12 & [-0.13, -0.11] & 0.47 (1.10) & -195.34*** & 0.56 & [0.55, 0.57] \\ 
& Zephyr & -1.99 (1.44) & -368.74*** & -1.62 & [-1.63, -1.61] & -1.27 (0.66) & -385.39*** & -2.10 & [-2.11, -2.09] \\ 

\bottomrule

\end{tabular}

\begin{tablenotes}
\small
\item \textit{Note.} $\Delta\mu$ = mean difference; $\sigma$ = standard deviation; WSR = Wilcoxon signed rank test; d = Cohen's d effect size; CI = confidence interval; ***p < .001.
\end{tablenotes}
\end{table*}

Figure \ref{fig:political-compass} illustrates the results of injecting ideological descriptors like ``right authoritarian" or ``left libertarian" to the personas descriptions, while Table \ref{tab:model-stats-combined} provides a detailed statistical analysis of the observed shifts.
When personas were prompted to adopt ``right-authoritarian" descriptor, all models exhibited significant vertical (authoritarian-libertarian) and horizontal (left-right) shifts. Notably, the shifts in the authoritarian dimension were consistently larger across most of the models. 
This suggests a higher susceptibility to ideological repositioning along the vertical axis compared to the horizontal axis. The only exception was Zephyr, which showed greater movement along the horizontal axis ($\Delta\mu_x$ = 1.60) compared to the vertical one ($\Delta\mu_y$ = 1.13).

Among the models, Llama displayed the most substantial overall movement ($\Delta\mu_x$ = 2.19, $\Delta\mu_y$ = 3.20), indicating that it is particularly sensitive to ideological manipulation. In contrast, Zephyr demonstrated the most resistance, with relatively modest changes ($\Delta\mu_x$ = 1.60, $\Delta\mu_y$ = 1.13). The pronounced vertical and horizontal movements align with the idea that right-authoritarian ideology contrasts sharply with the left-libertarian bias intrinsic to LLMs, as evidenced in their baseline distributions (Figure \ref{fig:combined}), thus facilitating more distinct repositioning relative to the original placement. 

For the ``left-libertarian" condition, the shifts were generally smaller in magnitude. Llama and Qwen showed minimal movement on the horizontal axis ($\Delta\mu_x = 0.16$ and $-0.15$, respectively) and moderate changes on the vertical axis ($\Delta\mu_y$ = 0.68 and 0.47). This discrepancy might stem from the models' pre-existing left-libertarian inclinations, making further adjustments less pronounced. This asymmetry may reflect biases in pretraining data, where left-libertarian narratives are more prevalent, leading to stronger internal representations of these ideologies.
Additionally, the  differences in responsiveness between left-right shifts versus authoritarian-libertarian shifts reveal that LLMs' internal conceptualization of political ideologies may not fully align with the political compass framework.

The effect sizes shown in Table \ref{tab:model-stats-combined} confirm the patterns observed in the mean shifts. For the ``right-authoritarian" condition, vertical shifts consistently exhibited large effect sizes across all models, with all the values exceeding 1.0 (e.g., Llama: d = 2.51; Qwen: d = 2.37). In contrast, horizontal shifts showed smaller but still significant effects (e.g., Mistral: d = 0.84). For the ``left-libertarian" condition, effect sizes were generally smaller, particularly for horizontal shifts, where values were near or below 0.20 for Llama and Qwen. 

Finally, when comparing the clustering versus spread of distributions after the descriptor injection, we find particularly interesting Zephyr's remarkable consistency across all three conditions reported in Table \ref{tab:avg-results}, with average distances from centroids remaining nearly constant (1.164, 1.175, and 1.163). This observation corroborates our previous finding that Zephyr demonstrated the strongest resistance to ideological shifts, as noted in the analysis of ``right-authoritarian" descriptor. In contrast, the other models showed more pronounced variations: Mistral exhibited substantial changes across conditions, Llama maintained relatively high dispersion throughout, while Qwen2.5 showed a trend toward increased clustering in both injection conditions.

\section{CONCLUSION}
In this study we investigated how adaptable LLMs are when impersonating profiles that may carry certain political ideologies.
Our results show that persona descriptions significantly influence LLMs' political compass placements, with models exhibiting varying degrees of malleability to both implicit persona characteristics and explicit ideological prompting. This has important implications for understanding and controlling their political biases.

\paragraph{\textbf{Limitations}} 
While the PCT provides valuable insights, it represents just one possible measure of political orientation. Alternative frameworks, like the 8values test, could provide additional dimensions beyond the economic and social axes of the PCT, offering a more nuanced political positioning across multiple value spectrums. Additionally, models' internal conceptualizations of political descriptors may not perfectly align with PCT definitions, potentially affecting our results' interpretation. While we recognize this limitation, our study's primary focus is not on evaluating whether language models accurately map personas to their real-world demographic political leanings. Instead, we aim to analyze broader patterns in the distribution of responses and examine the range of variability that different personas can elicit from these models.

\paragraph{\textbf{Future studies}} Looking forward, we identify three promising research directions: i) extending this analysis to larger models (70B+ parameters) could reveal how model scale affects political malleability; ii) deeper investigation of persona-specific biases could uncover important patterns in how models stereotype different roles and professions; and iii) these insights could inform techniques for de-biasing LLMs toward more neutral political orientations, potentially improving their utility across diverse applications.

\paragraph{\textbf{Acknowledgments}}
This work is partially supported by the Australian Research Council (ARC) Training Centre for Information Resilience (Grant No. IC200100022) and by an ARC Future Fellowship Project (Grant No. FT240100022).

\bibliographystyle{ACM-Reference-Format}
\bibliography{political-personas}

\end{document}